\pdfoutput=1

\documentclass[11pt]{article}

\usepackage[final]{acl}

\usepackage{times}
\usepackage{latexsym}
\usepackage{booktabs}

\usepackage[T1]{fontenc}

\usepackage[utf8]{inputenc}

\usepackage{microtype}

\usepackage{inconsolata}

\usepackage{graphicx}

\usepackage{amsmath}
\usepackage{amsfonts}
\usepackage{xspace}
\usepackage{makecell}
\usepackage{hyperref}
\usepackage{enumitem}
\usepackage{subcaption}
\usepackage{colortbl}

\usepackage{etoolbox}
\usepackage{mdframed}
\usepackage{multicol}
\usepackage{tabularx}
\usepackage{soul}
\usepackage[normalem]{ulem}
\usepackage{leftidx}
\definecolor{OliveGreen}{rgb}{0,0.6,0}
\definecolor{CornellRed}{rgb}{0.7, 0.11, 0.11}

\newmdenv[
  backgroundcolor=gray!5,
  linecolor=gray!40,
  linewidth=0.8pt,
  roundcorner=5pt,
  skipabove=12pt,
  skipbelow=12pt,
  innertopmargin=6pt,
  innerbottommargin=6pt,
  innerleftmargin=10pt,
  innerrightmargin=10pt,
]{promptbox}

\title{\projectname: Consistent Distractor Generation in Math MCQs}

\author{$\textbf{Nisarg Parikh}^1, \textbf{Alexander Scarlatos}^1$\thanks{Equal Contribution.}, $\textbf{Nigel Fernandez}^1$\footnotemark[1], \\ $\textbf{Simon Woodhead}^2, \textbf{Andrew Lan}^1$ \\
  $\text{University of Massachusetts Amherst}^1, \text{Eedi}^2$ \\
\texttt{\{nkparikh,nigel,ajscarlatos,andrewlan\}@cs.umass.edu} \\ 
  \texttt{simon.woodhead@eedi.co.uk} \\}

\newcommand{\as}[1]{\textcolor{blue}{\bf [alex: #1]}}

\newcommand{\projectname}[0]{\textsc{LookAlike}\xspace}
\newcommand{\lookalike}[0]{\textsc{LookAlike}\xspace}

\begin{document}
\maketitle

\setcounter{footnote}{2}

\begin{abstract}
Large language models (LLMs) are increasingly used to generate distractors for multiple-choice questions (MCQs), especially in domains like math education. 
However, existing approaches are limited in ensuring that the generated distractors are consistent with common student errors. 
We propose \projectname~
\footnote{Code: \href{https://github.com/umass-ml4ed/LookAlike}{https://github.com/umass-ml4ed/LookAlike}}
, a method that improves error–distractor consistency via preference optimization. Our two main innovations are: 
(a)~mining synthetic preference pairs from model inconsistencies, and 
(b)~alternating supervised fine-tuning (SFT) with Direct Preference Optimization (DPO) to stabilize training. Unlike prior work that relies on heuristics or manually annotated preference data, \projectname uses its own generation inconsistencies as dispreferred samples, thus enabling scalable and stable training. Evaluated on a real-world dataset of $1\text{,}400$+ math MCQs, \projectname achieves $51.6$\% accuracy in distractor generation and $57.2$\% in error generation under LLM-as-a-judge evaluation, outperforming an existing state-of-the-art method ($45.6$\% / $47.7$\%). These improvements highlight the effectiveness of preference-based regularization and inconsistency mining for generating consistent math MCQ distractors at scale. \end{abstract}

\section{Introduction}
\label{sec:introduction}
Multiple-choice questions (MCQs) are used in educational assessments \citep{anthony1996,peter2001,kubiszyn2016} to evaluate student understanding across various subjects and grades \cite{thomas2025does}. An MCQ consists of a question stem and a set of options, including a correct answer and multiple incorrect alternatives, referred to as \textit{distractors} \citep{fernandez2024divert,feng2024exploringad}. 
\textit{Distractors} are incorrect answers that students reach by making an \emph{error} while answering the question. It can be rooted in many ways, e.g., the student overgeneralizing to a new context, exhibiting an ingrained misconception, or simply slipping and being careless. Designing effective distractors can be crucial to the assessment and pedagogical aspects of MCQs \cite{simkin2005multiple}, since they help us identify student errors and prepare ways to mitigate them. 

Hand-crafting high-quality distractors requires extensive human effort by content designers and teachers since it requires them to anticipate common student errors, which can be difficult in subjects like math. Therefore, recent works have leveraged artificial intelligence, especially large language models (LLMs), to automate this process. Previous works on \textit{distractor} generation for MCQs have attempted to prompt LLMs to generate distractors~\cite{feng2024exploringad}, as well as fine-tune LLMs to generate possible student errors and then distractors caused by such errors, as shown in \mbox{DiVERT}~\cite{fernandez2024divert}. As noted in these works, the bottleneck in distractor generation performance is \emph{consistency}: LLMs are often capable of identifying mathematically feasible errors, but struggle at following such erroneous instructions to arrive at the corresponding distractor (a similar finding was also made in \cite{sonkar2024malalgoqa}). 
As shown in Table~\ref{tbl:consistent-examples}, both fine-tuned LLMs and the LLMs in \mbox{DiVERT} sometimes fail to follow the input error explanation to arrive at a consistent distractor. In the second example, the fine-tuned LLM fails to follow the error, ``finds $13\%$ of an amount rather than the percentage being asked'', arriving at an inconsistent distractor ($12$) rather than the consistent distractor ($5.2$).

\begin{table}
\small
\centering
\begin{tabular}{p{0.75\linewidth}p{0.1\linewidth}}

\toprule

\multicolumn{2}{c}{Question stem: Calculate: \( 130 \% \) of \( 40= \square \)}\\ 
\midrule
Error & Distractor\\
\midrule

\multicolumn{2}{c}{\textcolor{OliveGreen}{Plausible} error, \textcolor{OliveGreen}{plausible} and \textcolor{OliveGreen}{consistent} distractor.}\\
\midrule
Added the values together instead of finding the percentage. & $170$ \\

\midrule
\multicolumn{2}{c}{\textcolor{OliveGreen}{Plausible} error, \textcolor{OliveGreen}{plausible} but 
\textcolor{CornellRed}{inconsistent} distractor.}\\
\midrule
Finds 13\% of an amount rather than the percentage being asked. & $12$\\

\midrule
\multicolumn{2}{c}{\textcolor{CornellRed}{Implausible} error, \textcolor{OliveGreen}{plausible} but 
\textcolor{CornellRed}{inconsistent} distractor.}\\
\midrule
When solving a problem that requires an inverse operation (e.g. missing number problems), does the original operation.
 & $90$\\

\midrule
\multicolumn{2}{c}{\textcolor{CornellRed}{Implausible} error, \textcolor{CornellRed}{implausible} and 
\textcolor{CornellRed}{inconsistent} distractor.}\\
\midrule
Does not understand that 100\% is the whole amount.  & $20$\\

\bottomrule
\end{tabular}
\caption{Examples of inconsistent error-distractor pairs generated by SFT (second and fourth pairs), and a state-of-the-art method, DiVERT~\citep{fernandez2024divert} (third pair). \lookalike mines generation inconsistencies for scalable preference optimization. 
}
\label{tbl:consistent-examples}
\end{table}

To address this limitation, one natural solution is to regularize an LLM-based distractor generator, which takes the question stem and an error as input, to enforce that the generated distractor matches the input error. 
To this end, we resort to preference optimization, specifically direct preference optimization (DPO)~\cite{rafailov2023direct}. DPO training requires \emph{preference pairs} among outputs, i.e., a distractor that matches the input error and a distractor that does not. However, we empirically find two main challenges in using DPO to promote error-distractor consistency:

\begin{itemize}[noitemsep]
    \item Acquiring high-quality preference data typically requires costly manual annotation or unreliable synthetic heuristics \citep{li2023synthetic,tan2024largeannotation}, which is difficult due to the nature of the distractor generation task. 
    \item Models trained with DPO may deteriorate in quality after a few epochs~\citep{pal2024smaug,liu2024provably,yan2025dproperties,xu2024isdposuperior}, showing training instability.  
\end{itemize}

\subsection*{Contributions}

In this paper, we introduce \lookalike, proposing two methods to tackle these challenges and improve error-distractor consistency in math MCQs
For the first challenge, we create preference pairs by generating \textit{synthetic negative samples}: we evaluate LLM-generated errors, in addition to distractors, and use inconsistently generated errors and distractors as informative negative samples. 
This method creates meaningful signals that, when used in conjunction with consistent errors and distractors in DPO training, improve the consistency of LLMs in distractor generation. 
For the second challenge, we employ a regularization method in DPO training, which performs supervised finetuning (SFT) and DPO \emph{alternatively} in consecutive training iterations, which performs better than combining them both into a single objective, as done in recent works~\cite{liu2024provably,pal2024smaug}. 

We conduct extensive experiments on a real-world dataset containing math MCQs used by hundreds of thousands of students, with human-written error descriptions behind each distractor. 
Results show that \lookalike, compared to state-of-the-art baselines, improves distractor generation performance by up to 6\%. 
We also show that \lookalike improves error generation by up to 10\%, using an LLM-as-a-Judge evaluation. 
We also provide qualitative examples and an error analysis highlighting the improved consistency of generated \textit{errors} and \textit{distractors}. 

\section{Background}
\label{sec:background}
In this section we formally introduce the tasks of \textit{error} and \textit{distractor} generation in math MCQs. 
We also detail a baseline for preference pair creation and a baseline for DPO regularization, combining preference alignment with supervised learning.

\subsection{Task Definition}
\label{subsec:task-definition}
We consider an MCQ 
\( Q \) defined by its textual components: 
a \textbf{question stem} \( s \), (optionally) 
its \textbf{correct answer} or \textbf{key} \( k \), (optionally) an explanation of the key $f$, (optionally) question topic/concept tags $t$, 
and a set of \textbf{incorrect answer options} called  ground truth 
distractors $ D $.
Each $d_i \in D$ is (optionally) associated with a corresponding ground truth human-written \textbf{error explanation} or error $e_i \in E$.
All textual components above are represented as sequences of words and math symbols.
We aim to model the space of plausible student errors \( E \) and their corresponding distractors \( D \). We define two primary tasks:

\begin{enumerate}[leftmargin=0.5cm, noitemsep]
    \item \textbf{Error Generation:}  
    Learn an LLM parameterized model, $LLM^{err}(s,k,f,t,d_i) \rightarrow \hat{e}_i$, that outputs an error description $\hat{e}_i$ consistent with the given input distractor $d_i$ and MCQ.

    \item \textbf{Distractor Generation:}  
    Learn an LLM parameterized model, $LLM^{dis}(s,k,f,t,e_i) \rightarrow \hat{d}_i$, that outputs a distractor $\hat{d}_i$ consistent with the given error description $e_i$ and MCQ.
\end{enumerate}

\subsection{Baseline: Preference Pairs from Ground-truth Error-Distractor Pairs}
\label{subsec:preference-pairs-from-ground-truth}
As a natural starting point, following a similar method from \cite{scarlatos2024improving}, one can construct preference pairs for DPO as follows: For each question, there are multiple distractors ($D= {d_1, d_2, \dots, d_n}$) and their corresponding errors ($E = {e_1, e_2, \dots, e_n}$). As a baseline, for the error behind the $i^{th}$ distractor, $e_i$, we can use $d_i$ itself as the preferred response, and use the remaining distractors ($d_j \in D \setminus \{d_i\}$) as dispreferred responses. We use a similar procedure for the errors.
We dub this method for preference pair construction as \textbf{DPO-GT} (ground truth).
However, the number of dispreferred responses is limited by the number of human-written error-distractor pairs for the question. \lookalike, on the other hand, creates preference pairs by generating \textit{synthetic negative samples}, allowing for an arbitrary number of dispreferred responses for scalable preference optimization, resulting in improved consistency in both error and distractor generation (Section~\ref{subsec:mining_preference_pairs}).

\subsection{Baseline: DPO Regularization}
\label{subsec:optimizing-dpo-sft-objectives-jointly}
Models trained with DPO have been shown to deteriorate in quality after a few epochs due to training instability~\citep{pal2024smaug,liu2024provably,yan2025dproperties,xu2024isdposuperior}. Existing regularization techniques to improve DPO training stability include Regularized Preference Optimization (\textbf{RPO})~\cite{liu2024provably}, and DPO-Positive (\textbf{DPOP})~\cite{pal2024smaug}. RPO optimizes both the DPO loss and the SFT loss jointly, i.e., $L_{RPO} = L_{DPO} + \lambda \beta L_{SFT}$. 
The SFT loss uses the preferred response as the ground-truth completion. RPO suffers from conflicting gradient directions \citep{shi2023recon, liu2024conflictaversegradientdescentmultitask}, especially when the preference-based signal (DPO) incentivizes ranking decisions that are misaligned with the next-token prediction signal (SFT). 
DPOP uses the SFT objective as a penalty but their improvement is limited to preference pairs with high edit distances between them. 
\lookalike, on the other hand, proposes an alternating optimization approach to stabilize DPO training, interleaving SFT and DPO training either at the per-batch or per-epoch level, resulting in improved consistency in both error and distractor generation compared to RPO and DPOP (Section~\ref{subsec:alternating_sft_dpo_introduction}).

\section{Methodology}
\label{sec:methodology}
\begin{figure*}[!ht]
  \centering
  \includegraphics[width=0.9\linewidth]{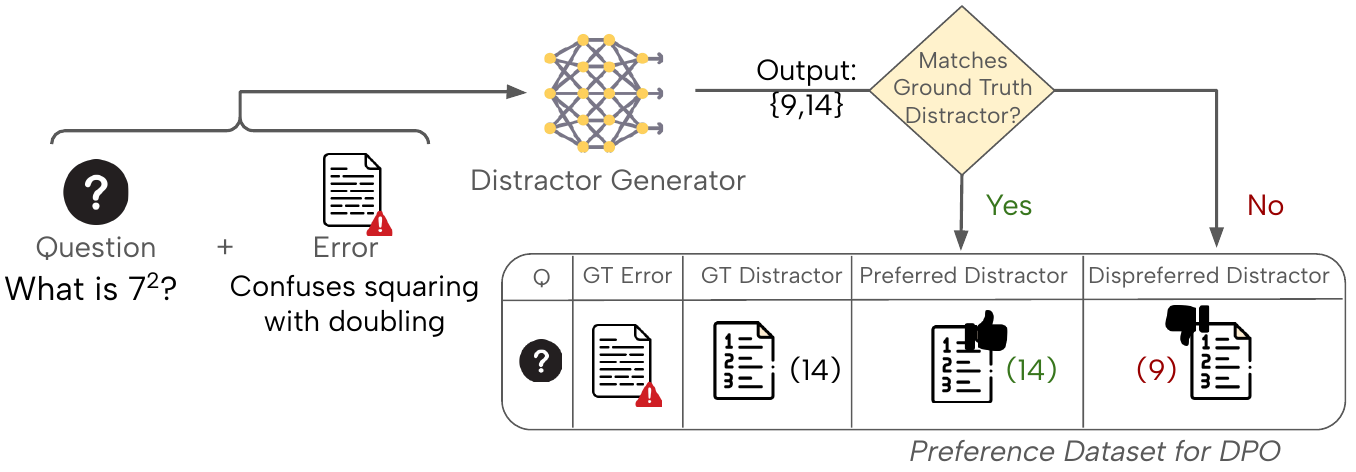}
  \captionof{figure}{
  \lookalike creates preference pairs by overgenerating a set of distractors for a question and error, and preferring those that match the ground-truth distractor exactly. An analogous process for error generation. 
  }
  \label{fig:synthetic-dataset-generation}
\end{figure*}

\begin{figure}[!ht]
  \centering
  \includegraphics[width=\linewidth]{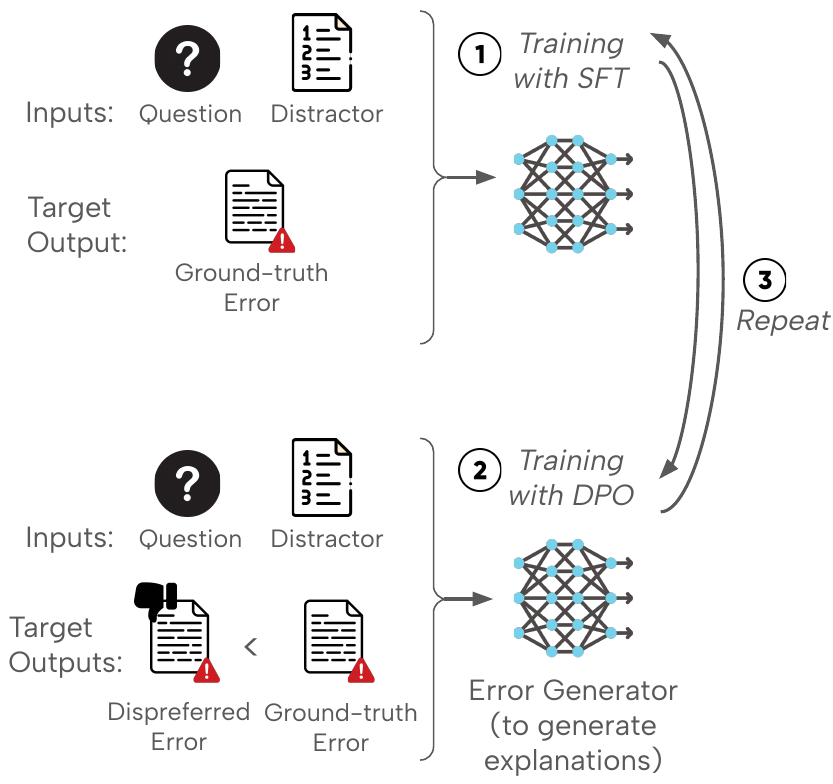}
  \captionof{figure}{
  \lookalike employs an alternating optimization strategy, switching between SFT and DPO objectives to regularize DPO training.
  }
  \label{fig:alternating-regularization}
\end{figure}

We now detail our framework, \lookalike, which a) creates preference pairs by generating synthetic negative samples, and b) employs a DPO regularization technique of alternating optimization between SFT and DPO for better training stability, leading to improved error and distractor generation consistency.

\subsection{Mining Preference Pairs via Inconsistencies for DPO}
\label{subsec:mining_preference_pairs}
Prior work~\cite{fernandez2024divert} has highlighted a significant issue of consistency in distractor generation performance, with LLMs struggling to follow error descriptions to arrive at corresponding distractors, examples of which are shown in Table~\ref{tbl:consistent-examples}. \lookalike mines these generation inconsistencies as \textit{synthetic negative samples} to create preference pairs for DPO training.

We visualize our preference pair creation in \lookalike in Figure~\ref{fig:synthetic-dataset-generation}. For distractor generation, $LLM^{dis}$ overgenerates a set of distractors for an input question stem and a ground-truth error. Each generated distractor is then compared against the ground-truth distractor.
In our preference dataset, generated distractors that match the ground-truth distractor exactly are preferred responses, while those that do not exactly match the ground-truth distractor are dispreferred responses. A similar process is applied to create preference pairs for error generation, with exact string match\footnote{LLM-as-a-Judge using GPT-4o-mini as a similarity measure led to lower performance.} 
used to compare generated errors against the ground-truth error to form preference pairs.

Formally, given an MCQ dataset with samples, $(s, e, d)$, where s is the question stem, $e$ is the error description, and $d$ is the corresponding distractor, we first train a distractor generation model, $LLM^{dis}$, to output the corresponding distractor through SFT. To create preference pairs, we then overgenerate multiple distractors $\hat{d} \in \hat{D}$ from the fine-tuned $LLM^{dis}$ for each $(s,e)$ pair. For each generated distractor $\hat{d}$, we check if $\hat{d}$ matches the ground-truth distractor $d$ exactly. If yes, we add $\hat{d}$ as a preferred response, and if no, we add $\hat{d}$ as a dispreferred response in our distractor generation preference dataset. 
Having constructed the preference dataset, we further train our fine-tuned $LLM^{dis}$ through DPO~\cite{rafailov2023direct}. A similar process is applied to form our error generation preference dataset which is then applied for DPO training of $LLM^{err}$. Creating preference pairs from the static ground-truth dataset is limited by the number of human-written annotations (Section~\ref{subsec:preference-pairs-from-ground-truth}). \lookalike, on the other hand, uses generations from the currently fine-tuned LLM to create an arbitrary number of dynamic preference pairs, with negative preference signals being more aligned with the inconsistency failure modes of the fine-tuned LLM. 

\subsection{DPO Regularization Through Alternating Optimization}
\label{subsec:alternating_sft_dpo_introduction}

\begin{table*}
\small
\centering
\begin{tabular}{l|l}
\toprule
Question & \begin{tabular}[c]{@{}l@{}}\( \frac{3}{7} \) of a group of students are boys. What would be a possible ratio of boys to girls?\end{tabular} \\
\midrule
Key &  $3: 4$  \\
\midrule
Ground-truth Distractor & $3:10$ \\
\midrule
Ground-truth Error & Uses the denominator when converting from fractions to ratio, rather than numerator.\\
\midrule
Generated Error (Epoch 1) & Includes the denominator when converting a fraction to a ratio. \\
\midrule
Generated Error (Epoch 2) & \begin{tabular}[c]{@{}l@{}}When converting a fraction to a ratio, puts the other side of the ratio as the denominator.\end{tabular} \\
\midrule
Generated Error (Epoch 3) & \begin{tabular}[c]{@{}l@{}}When converting a fraction to a ratio, thinks you just use the numerator and denominator\\as the numbers in the ratio. Additionally, thinks you can use the denominator on its own\\as the total number of parts in a ratio.\end{tabular}\\
\bottomrule
\end{tabular}
\caption{Error generation quality deteriorates over DPO training epochs without using regularization. 
}
\label{tbl:per_epoch_OOD_examples}
\vspace{-0.1cm}
\end{table*}

We empirically observe that models trained with DPO deteriorate in quality after a few epochs due to training instability.
We show examples of degradation in error generation quality over three training epochs in Table~\ref{tbl:per_epoch_OOD_examples}. We observe errors become more verbose with an increase in length and are out-of-distribution from the human-written errors as the number of DPO training epochs increases, as also shown in prior work~\cite{park2024disentanglinglengthqualitydirect}.

To mitigate this issue, we introduce a regularization strategy that trains the error/distractor-generation LLM by \textit{alternating optimization}, i.e., by switching between 
SFT
and DPO
objectives during training, as shown in Figure~\ref{fig:alternating-regularization}. 
This alternating optimization allows the LLM to periodically recalibrate to the ground-truth distribution (via SFT) while remaining faithful to learning ranking preferences of consistent generations (via DPO). After each SFT optimization, the preference dataset is recomputed 
(Section~\ref{subsec:mining_preference_pairs}) for the subsequent DPO optimization, using the currently trained LLM for better alignment, allowing for dynamic and scalable preference pair creation. 
We experiment with alternating between SFT and DPO optimization at two different levels: per-batch and per-epoch, picking the one giving better performance empirically. For both levels, the preference dataset is recomputed after every epoch.

\paragraph{Alternating Optimization Per-Batch.} 
At each training step $t$, the LLM parameters $\theta$ are updated using a learning rate of $\eta$ following:
\begin{equation}
    \theta_{t+1} = \theta_t - \eta \nabla L(\theta_t),
\end{equation}
where the loss function $L$ alternates based on a batch-level schedule:
\begin{equation}
     L(\theta_t) =
     \begin{cases}
         L_{SFT}(\theta_t), & \text{if batch } t \text{ is even} \\
         L_{DPO}(\theta_t), & \text{otherwise}
     \end{cases}
 \end{equation}

\paragraph{Alternating Optimization Per-Epoch.}
As a coarser alternative, the loss function $L$ alternates based on an epoch-level schedule:
 \begin{equation}
     L(\theta_t) =
     \begin{cases}
        L_{SFT}(\theta_t), & \text{if epoch } t \text{ is even} \\
        L_{DPO}(\theta_t), & \text{otherwise}
     \end{cases}
 \end{equation}

\section{Experimental Evaluation}
\label{sec:experimental_evaluation}

In this section, we detail our experiments on a real-world math MCQ dataset, evaluating the efficacy of \lookalike in comparison with state-of-the-art baselines for both distractor generation and error generation.

\subsection{Dataset}
\label{subsec:tasks_and_dataset}
We conduct our experiments on a real-world math MCQ dataset from a large learning platform used by hundreds of thousands of students. The dataset consists of $1,434$ math MCQs, each containing a question stem, key, explanation of the key, topic/concept tags, and $3$ distractors along with their respective teacher-written error descriptions explaining why a student might select that distractor. The MCQs are designed for students aged between $10$ to $13$ and span $41$ distinct mathematical subtopics, including \textit{Arithmetic}, \textit{Fractions}, and \textit{Solving Equations}. We split the dataset into training, validation, and test by questions to ensure no overlap across splits using a $72$\%-$16$\%-$12$\% proportion. See Appendix~\ref{adx:examples-from-eedi}
for math MCQ examples.

\subsection{Baselines}
\label{subsec:baselines}
We compare \projectname with $3$ baselines. The \textbf{SFT} baseline, used as a baseline in \cite{fernandez2024divert}, fine-tunes an LLM to generate the corresponding distractor (or error) given the question and the error (or distractor) as input. 
The \textbf{\mbox{DiVERT}}~\cite{fernandez2024divert} baseline employs a variational approach to learn an interpretable error space behind distractors. Post variational training, we use the error generation and distractor generation LLMs from DiVERT as baselines.
We also compare against forming preference pairs from the ground-truth error-distractor pairs; we continue training the SFT baseline on this preference dataset using DPO and refer to the resulting model as \textbf{DPO-GT} (Section~\ref{subsec:preference-pairs-from-ground-truth}).
For fairness, we regularize DPO training for DPO-GT by exploring all techniques (RPO, DPOP, our alternating per-batch optimization, and our alternating per-epoch optimization), and choose the regularization (per-epoch) that results in the best performance.

\subsection{Metrics}
\label{subsec:metrics_and_baselines}

\paragraph{Distractor Evaluation.}
\label{subsubsec:distractor_generation}
Following prior work on distractor generation~\cite{fernandez2024divert,feng2024exploringad}, we use \textbf{Exact match} as our evaluation metric to measure alignment between the generated distractor and the ground-truth distractor corresponding to a question and error.

\paragraph{Error Evaluation.}
\label{subsubsec:error_generation}

Automated text similarity metrics like exact string match, ROUGE-L F1~\cite{lin2004rouge}, or BERTScore F1~\cite{zhang2020bertscoreevaluatingtextgeneration} are unsuitable for error evaluation given the open-ended and mathematical nature of errors. We therefore adopt an \textbf{LLM-as-Judge}~\cite{liu2023geval,zheng2023judging} evaluation, prompting GPT-4o-mini to evaluate if the generated error is mathematically equivalent 
to the ground-truth error given the question and corresponding distractor. We show our prompt in Appendix~\ref{adx:llm-as-a-judge}. 

\subsection{Implementation Details}
\label{sec:implementation_details}
Following prior work~\cite{fernandez2024divert}, all methods use MetaMath-Mistral $7$B~\cite{yu2024metamathbootstrapmathematicalquestions} as their base LLM, as we found it provides a suitable prior within the $7$B parameter size models for mathematical reasoning. At test time, we use standard beam search with $10$ beams for distractor generation, and diverse beam search~\cite{vijayakumar2018diverse} with $10$ beams for error generation. Detailed hyperparameter settings for all methods are provided in Appendix~\ref{adx:baselines}.

To ensure fair comparison, we limit \lookalike’s synthetic generation to 3 distractors and 3 errors per training sample per epoch, resulting in a similar order of magnitude of training samples as DPO-GT. We also use the same training budget and regularization for both methods. All fine-tuned models, including SFT and DPO-based variants, were trained with LoRA to ensure parameter efficiency and consistency in comparison.  

\section{Results, Analysis and Discussion}
\label{sec:results_and_discussion}
In this section, we detail our experimental results. We quantitatively evaluate the quality of generated errors and distractors, qualitatively evaluate the consistency of generated errors through human evaluation, conduct an ablation study on DPO regularization techniques, and perform an error analysis on failed cases of error generation.  

\subsection{Quantitative Evaluation}
\label{sec:performance-of-lookalike}
Table~\ref{tbl:dpo_vs_sft} shows the average performance on distractor generation and error generation, across $5$ cross-validation folds, for all methods. DPO-based methods, DPO-GT and \lookalike, are trained using our alternating optimization technique for DPO regularization, choosing the alternating level (per-batch or per-epoch) that works best for downstream task performance. DPO-GT works best with per-epoch for both tasks, while \lookalike works best with per-epoch for distractor generation, and per-batch for error generation.

\paragraph{Preference optimization using inconsistent error-distractor pairs improves consistency.}
\begin{table}[!t]
\centering
\footnotesize
\begin{tabular}{lcc}
\toprule
 & \makecell{Distractor Gen\\ (Exact Match $\uparrow$)} & \makecell{Error Gen\\(LLM-as-Judge $\uparrow$)} \\
 \midrule
SFT & $44.76$ & $46.68$ \\
DiVERT & $45.64$ & $47.72$ \\
DPO-GT 
& $51.44$ & $57.02$ \\
\projectname & $\mathbf{51.56}$ & $\mathbf{57.18}$\\
\bottomrule
\end{tabular}
\caption{
Cross-validation performance on distractor generation and error generation for all methods across 5 folds. \lookalike 
outperforms SFT and the prior state-of-the-art method DiVERT~\cite{fernandez2024divert}, and is comparable to DPO-GT.
}
\label{tbl:dpo_vs_sft}
\end{table}

\lookalike outperforms SFT and the previous state-of-the-art baseline DiVERT~\cite{fernandez2024divert}, by a wide margin of $6.8\%$ and $5.92\%$ on distractor generation, and $10.5\%$ and $9.46\%$ on error generation performance, respectively. The improvement is statistically significant with $\text{p-values} < 0.05$ measured using a one-sample Wilcoxon signed-rank test~\cite{Rey2011wilcoxonsigned}. This result validates our idea of mining error-distractor inconsistencies as preference pairs for DPO training to improve both error and distractor generation consistency. Further, \lookalike, although using synthetic negative samples drawn from its own inconsistent generations as preference pairs, is comparable in performance to DPO-GT, which uses human-written annotations as preference pairs, demonstrating the potential and flexibility of \lookalike for scalable, domain-agnostic preference optimization.

Although the performance difference between \lookalike and DPO-GT appears small ($0.12\%$ and $0.16\%$ on distractor and error generation respectively), it is important to note that \lookalike achieves this using automatically mined preference pairs from inconsistent generations, without relying on ground-truth labels, highlighting its scalability. Moreover, the improvement over DiVERT ($5.9$-$10.5\%$) is substantial and statistically significant.

\paragraph{Alternating optimization is an effective DPO regularization.}
\begin{table}[!t]
\centering
\small
\begin{tabular}{lcc}
\toprule
 & \makecell{Dis Gen\\(Exact M. $\uparrow$)} & \makecell{Error Gen\\(LLM-as-Judge $\uparrow$)} \\ 
\midrule
\begin{tabular}[c]{@{}l@{}}DPO-GT w/o Reg.\end{tabular}
& $47.68$ & $53.96$ \\
+ DPOP & $47.80$ & $52.74$ \\
+ RPO & $49.14$ & $52.44$ \\ 
+ Per-batch
& $49.66$ & $55.74$ \\
+ Per-epoch 
& $51.44$ & $57.02$ \\
\midrule
\begin{tabular}[c]{@{}l@{}}\projectname w/o Reg.\end{tabular} & $47.98$ & $49.34$ \\
+ DPOP & $49.38$ & $49.44$ \\+ RPO & $49.60$ & $49.66$ \\ 
 + Per-batch & $50.84$ & $\mathbf{57.18}$ \\
+ Per-epoch & $\mathbf{51.56}$ & $56.64$ \\\bottomrule
\end{tabular}
\caption{
Ablation study of various DPO regularization techniques. Our alternating (per-batch/epoch) optimization performs best for both DPO-GT and \lookalike.
}
\label{tbl:compare_with_dpo_from_gt}
\end{table}

Table~\ref{tbl:compare_with_dpo_from_gt} shows an ablation study comparing different DPO regularization techniques to combat deterioration in generation quality~\cite{pal2024smaug} during DPO training. 
Existing approaches like DPOP~\cite{pal2024smaug} and RPO~\cite{liu2024provably} provide marginal gains up to $1.62\%$ for distractor generation and $0.32\%$ for error generation. Our alternation optimization, switching between SFT and DPO objective, at either the per-batch or per-epoch level, leads to the best performance for both, DPO-GT and \lookalike, with performance gains up to $1.96\%$ on the distractor generation task and $7.52\%$ on the error generation task. These results show that alternating optimization effectively guides the LLM to periodically recalibrate to the ground-truth distribution (via SFT) while remaining faithful to learning ranking preferences of consistent generations (via DPO).

\subsection{Qualitative Case Studies}
\label{sec:qualitative_analysis}
\paragraph{\lookalike generates more consistent errors.} 
Table~\ref{tbl:qualitative_analysis} shows errors from \lookalike compared to errors generated from SFT on two math questions. For the question on finding factors, SFT generates an overly generalized error applicable to many potential distractors, ``Does not understand the term factor''. On the other hand, \lookalike generates a more specific error, ``When asked for factors of an algebraic expression, thinks any part of a term will be a factor'', consistent with the distractor.
Similarly, for the question on simplifying algebraic terms, SFT generates an abstract error applicable to many distractors, ``Tries to add or subtract unlike terms''. On the other hand, \lookalike generates a more specific and consistent error leading to the input distractor, ``When collecting like terms, treats subtractions as if they are additions.''
We see similar patterns across other topics, with errors generated by \lookalike being more specific and consistent with the input question and distractor.
We also show qualitative examples of generated errors across all methods in Appendix~\ref{adx:comparing-errors-across-all}.

\paragraph{Error Analysis of \projectname.}
While \projectname outperforms SFT in generating more consistent errors and distractors, we observe some examples of generated errors that are inconsistent with the input question-distractor pair. One failure pattern observed is of \textit{template overfitting}, where \lookalike generates an error by overfitting to the error-distractor template of a similar question seen during training, generating errors that are consistent with other distractors from similar questions but not the input distractor. 
Table~\ref{tab:appendix-error-comparison} in the Appendix shows two examples. We see that the generated error, ``Has multiplied by the root power'', is inconsistent with the input distractor $64$, but upon inspection, is present as a ground-truth error and consistent with another question-distractor pair on the same topic.

\begin{table*}[]
\centering
\small
\begin{tabular}
{p{0.2\linewidth}|p{0.38\linewidth}p{0.33\linewidth}}
\toprule 

Topic & Finding factors & Simplifying terms \\
\midrule

Question Stem & Which of the following is a factor of: $6 n^{2}-9$? & 
Simplify the following expression by collecting like terms: $6 x-2 y-x+3 y$. \\
\midrule

Key & $3$ & $5 x+y$ \\
\midrule

Ground-truth Distractor & 9 & $7x+5y$ \\
\midrule

Ground-truth Error & When asked for factors of an algebraic expression, thinks a term will be a factor. & 
When collecting like terms, treats subtractions as if they are additions.\\
\midrule

SFT-Generated Error & Does not understand the term factor. & Tries to add or subtract unlike terms. \\
\midrule

\lookalike-Generated Error & When asked for factors of an algebraic expression, thinks any part of a term will be a factor. & When collecting like terms, treats subtractions as if they are additions.\\ 

\bottomrule
\end{tabular}
\caption{
Examples showing errors generated from \lookalike are more consistent than errors generated by SFT. 
}
\label{tbl:qualitative_analysis}
\end{table*}

\subsection{Human Evaluation}
\label{sec:human_analysis}
\begin{table}
\centering
\small
\begin{tabular}{cccc}
\toprule
 & Human & SFT & \lookalike \\ \midrule
Avg. Rating & 0.812 & 0.400 & 0.587 \\ \bottomrule
\end{tabular}
\caption{
Average error consistency rating by human evaluators. \lookalike generates more consistent errors than SFT.
}
\label{tbl:human-analysis}
\end{table}

\paragraph{Setup.}
We conduct a human evaluation on the quality and consistency of generated errors. We instruct two independent annotators with teaching experience to evaluate whether an error is consistent with a given input math question and corresponding distractor, choosing between a) yes, b) partially, and c) no.
Our instructions to human annotators are provided in Appendix~\ref{adx:human_analysis_instructions}.

We randomly select $40$ math questions from our test set spanning a diverse range of topics. For each question, we include its ground-truth human-written error, the error generated by SFT, and the error generated by \lookalike, for human evaluation. This process results in $120$ errors, along with their corresponding questions and distractors, for human evaluation. We shuffle the $120$ samples to avoid annotator bias. 

\paragraph{Results.}
Table~\ref{tbl:human-analysis} shows the average of annotators’ ordinal ratings on error explanations from the ground truth, SFT, and \projectname models. Ground truth errors scored the highest (mean = 0.812), followed by \projectname (0.587), and SFT (0.400). While \projectname does not match the human-authored ground truth, it significantly outperforms SFT on average, suggesting that preference-based regularization leads to more pedagogically consistent explanations.

We also measured agreement between annotators using quadratic-weighted Cohen’s kappa, and found that error labels generated by \projectname led to the highest agreement (0.740), surpassing both SFT (0.659) and even the inter-annotator agreement on ground truth labels (0.415). This result suggests that errors generated by \projectname are easier for humans to interpret consistently, even if they are not always as plausible as ground truth explanations. We see a lower agreement on ground truth errors because their pedagogical nuance and potential generality made consistency judgments more subjective for annotators compared to the often more literal AI-generated errors.

Finally, we compared agreement between evaluations from human annotators to evaluations from GPT-4o-mini-based LLM-as-Judge, our reference metric for error generation.  
Agreement varied between annotators, with the first annotator showing moderate agreement (linear Kappa) with GPT-4o-mini ($0.556$ for \lookalike-generated errors and $0.505$ for SFT-generated errors), and the second annotator showing low agreement ($0.314$ for \lookalike-generated errors and $0.409$ for SFT-generated errors). 

\section{Related Work}
\label{sec:related-work}
\paragraph{Error-Distractor Generation for Math MCQs.} 
Automated generation of math MCQs, and particularly their distractors, has progressed from template-based (rule-based and constraint-based) methods~\cite{shin2019topicmodelling,liang2018distractorrank,luo2024cotdistractor} to Large Language Model (LLM) approaches~\cite{fernandez2024divert,feng2024exploringad,scarlatos-etal-2024-improving,bitew2023predictiveprompting,chung2020bertdisgen}. 
A critical challenge, however, remains the generation of high-quality distractors that accurately reflect common student errors and misconceptions~\cite{alhazmi2024mcqsurvey,stasaski2017multipleontology}. 
Current methods advance error representation using variational techniques~\cite{fernandez2024divert}, RAG-based methods~\cite{yu2024enhancingrag}, and knowledge-bases~\cite{zhu2021cloze}. 

\paragraph{Preference Optimization in Education.} 
Preference learning techniques, including Reinforcement Learning from Human Feedback (RLHF) \cite{ouyang2022traininglanguagemodelsfollow} and its more stable, computationally efficient alternative Direct Preference Optimization (DPO)~\cite{rafailov2023direct}, are vital for aligning AI outputs with human judgments in education \cite{mon2024rlined}. 
Many recent approaches have used DPO~\cite{lee2025dpomcqgen,sonkar2024pedagogicalalignment,learnlmteam2024learnlmimprovinggeminilearning,ashok-kumar-lan-2024-improving,scarlatos2024improving,scarlatos2025trainingllmbasedtutorsimprove} but they do not handle some known failure modes of DPO related to inconsistent or out-of-distribution generation which the synthetic data generation of \projectname utilizes and the regularization of \projectname addresses. 
Other works mitigate these issues by providing regularization by using entropy \cite{shekhar2024seedposelfentropyenhanced}, length-based rewards  \cite{park2024disentanglinglengthqualitydirect}, or the SFT objective \cite{liu2024provably,pal2024smaug}, \projectname improved on these by providing a simpler SFT-based regularization approach which requires less hyperparameter tuning and is easier to apply. 

\paragraph{Challenges in Erroneous Instruction Following.}
Generating distractors from error descriptions, is an instance of the broader challenge of AI instruction following\cite{lou2024largelanguagemodelinstruction}. AI systems, including LLMs, struggle with complex reasoning \cite{heo2024douncertainity,son2024multitaskinference}, multi-step tasks \cite{chen2024sifo,wang2023learningmultistepreasoningsolving,fujisawa2024procbenchbenchmarkmultistepreasoning}, and adhering to multiple constraints simultaneously \cite{wen2024benchmarkingcomplexinstructionfollowing}, sometimes exhibiting a "curse of instructions" where performance degrades as complexity increases \cite{jang2022can,son2024multitaskinference}. Generalization also poses a significant hurdle; models often fail to apply instructions to new tasks or in novel combinations (compositional generalization) \cite{cohen2025compositionalinstructionfollowinglanguage,dan2021compositionaldata}. These challenges can lead to inconsistencies where the generated output does not faithfully reflect the nuances of the input instruction \cite{jang2022can,son2024multitaskinference,heo2024douncertainity}, a problem \lookalike aims to mitigate in the context of error-distractor generation through targeted preference optimization.

\section{Conclusion}
\label{sec:conclusion}

In this paper, we introduced \lookalike, a method that improves error-distractor consistency in math MCQs via preference optimization. 
\lookalike uses two main innovations: a) mining synthetic preference pairs from model generation inconsistencies and b) alternating optimization by switching between SFT and DPO objectives to stabilize training. 
Through extensive experiments on a real-world math MCQ dataset, we showed that \projectname outperforms the previous state-of-the-art method by a wide margin on both error generation and distractor generation.
These improvements highlighted the potential of inconsistency mining and preference-based regularization for generating consistent math MCQ distractors at scale.
We identify several limitations and avenues for future work. First, while \lookalike improves error and distractor generation consistency, examples of inconsistent generations remain. Ideas for creating preference pairs using error generation and distractor generation models together could be a promising direction. Second, testing the generalizability of \lookalike to math MCQs from unseen topics remains unexplored.

\clearpage

\section*{Limitations}
\label{sec:limitations}

While \projectname demonstrates improvements in generating consistent error-distractor pairs, it currently operates within the domain of middle-school mathematics. Extending the approach to other subjects like science or language arts may require minor modifications to the error and distractor representations. 

Additionally, the current preference mining strategy relies on model-generated inconsistencies, which assumes the base model is sufficiently trained to surface pedagogically meaningful contrastive samples. 
In practice, we find that models pretrained on math data (e.g., MetaMath) meet this assumption, suggesting this is a broadly applicable approach rather than a bottleneck.

Our use of exact match to label non-matching outputs as dispreferred is conservative and intentionally strict; it helps emphasize high-confidence inconsistencies. Nonetheless, exploring softer similarity-based criteria or human judgments to refine preference mining is a valuable future direction.

\section*{Ethical Considerations}
\label{sec:ethics_considerations}
Our goal is to reduce educator workload by automating the generation of plausible distractors and their associated misconceptions, ultimately supporting teachers in providing more personalized feedback. 
However, we acknowledge a potential concern around over-reliance on AI-generated content in educational settings. 
While our system is designed to assist, not replace, educators, thoughtful deployment practices and educator-in-the-loop designs are encouraged.

The use of large language models (LLMs) introduces the standard risks of inherited biases or artifacts from pretraining data. 
In our case, these risks are minimal, as the domain of application (mathematical misconceptions) is highly constrained and less prone to sociolinguistic biases.
Nevertheless, we encourage ongoing validation and periodic audits as best practices when deploying AI systems in learning environments.

\section*{Acknowledgments}
This work is partially supported by Renaissance Philanthropy via the learning engineering virtual institute (LEVI) and NSF grants 2118706, 2237676, and 2341948. We thank Hasnain Heickal and Zhangqi Duan for helpful discussions and annotations regarding this work.


\bibliography{custom}

\begin{thebibliography}{57}
\providecommand{\natexlab}[1]{#1}

\bibitem[{Airasian(2001)}]{peter2001}
Peter Airasian. 2001.
\newblock Classroom assessment: Concepts and applications.
\newblock \emph{McGraw-Hill, Ohio, USA}.

\bibitem[{Alhazmi et~al.(2024)Alhazmi, Sheng, Zhang, Zaib, and Alhazmi}]{alhazmi2024mcqsurvey}
Elaf Alhazmi, Quan~Z. Sheng, Wei~Emma Zhang, Munazza Zaib, and Ahoud Alhazmi. 2024.
\newblock Distractor generation in multiple-choice tasks: A survey of methods, datasets, and evaluation.

\bibitem[{Ashok~Kumar and Lan(2024)}]{ashok-kumar-lan-2024-improving}
Nischal Ashok~Kumar and Andrew Lan. 2024.
\newblock \href {https://aclanthology.org/2024.bea-1.10/} {Improving socratic question generation using data augmentation and preference optimization}.
\newblock In \emph{Proceedings of the 19th Workshop on Innovative Use of NLP for Building Educational Applications (BEA 2024)}, pages 108--118, Mexico City, Mexico. Association for Computational Linguistics.

\bibitem[{Bitew et~al.(2023)Bitew, Deleu, Develder, and Demeester}]{bitew2023predictiveprompting}
Semere~Kiros Bitew, Johannes Deleu, Chris Develder, and Thomas Demeester. 2023.
\newblock Distractor generation for multiple-choice questions with predictive prompting and large language models.

\bibitem[{Chen et~al.(2024)Chen, Liao, Qi, Eustratiadis, Monz, Bisazza, and de~Rijke}]{chen2024sifo}
Xinyi Chen, Baohao Liao, Jirui Qi, Panagiotis Eustratiadis, Christof Monz, Arianna Bisazza, and Maarten de~Rijke. 2024.
\newblock The {SIF}o benchmark: Investigating the sequential instruction following ability of large language models.
\newblock In \emph{Findings of the Association for Computational Linguistics: EMNLP 2024}. Association for Computational Linguistics.

\bibitem[{Chung et~al.(2020)Chung, Chan, and Fan}]{chung2020bertdisgen}
Ho-Lam Chung, Ying-Hong Chan, and Yao-Chung Fan. 2020.
\newblock A {BERT}-based distractor generation scheme with multi-tasking and negative answer training strategies.
\newblock In \emph{Findings of the Association for Computational Linguistics: EMNLP 2020}. Association for Computational Linguistics.

\bibitem[{Cohen et~al.(2025)Cohen, Tasse, Gopalan, James, Gombolay, Mooney, and Rosman}]{cohen2025compositionalinstructionfollowinglanguage}
Vanya Cohen, Geraud~Nangue Tasse, Nakul Gopalan, Steven James, Matthew Gombolay, Ray Mooney, and Benjamin Rosman. 2025.
\newblock Compositional instruction following with language models and reinforcement learning.

\bibitem[{Dan et~al.(2021)Dan, Han, and Roth}]{dan2021compositionaldata}
Soham Dan, Xinran Han, and Dan Roth. 2021.
\newblock Compositional data and task augmentation for instruction following.
\newblock In \emph{Findings of the Association for Computational Linguistics: EMNLP 2021}. Association for Computational Linguistics.

\bibitem[{Fahad~Mon et~al.(2023)Fahad~Mon, Wasfi, Hayajneh, Slim, and Abu~Ali}]{mon2024rlined}
Bisni Fahad~Mon, Asma Wasfi, Mohammad Hayajneh, Ahmad Slim, and Najah Abu~Ali. 2023.
\newblock Reinforcement learning in education: A literature review.
\newblock \emph{Informatics}.

\bibitem[{Feng et~al.(2024)Feng, Lee, McNichols, Scarlatos, Smith, Woodhead, Ornelas, and Lan}]{feng2024exploringad}
Wanyong Feng, Jaewook Lee, Hunter McNichols, Alexander Scarlatos, Digory Smith, Simon Woodhead, Nancy Ornelas, and Andrew Lan. 2024.
\newblock Exploring automated distractor generation for math multiple-choice questions via large language models.
\newblock In \emph{Findings of the Association for Computational Linguistics: NAACL 2024}. Association for Computational Linguistics.

\bibitem[{Fernandez et~al.(2024)Fernandez, Scarlatos, Feng, Woodhead, and Lan}]{fernandez2024divert}
Nigel Fernandez, Alexander Scarlatos, Wanyong Feng, Simon Woodhead, and Andrew Lan. 2024.
\newblock {D}i{VERT}: Distractor generation with variational errors represented as text for math multiple-choice questions.
\newblock In \emph{Proceedings of the 2024 Conference on Empirical Methods in Natural Language Processing}. Association for Computational Linguistics.

\bibitem[{Fujisawa et~al.(2024)Fujisawa, Nobe, Seto, Onda, Uchida, Ikoma, Chien, and Kanai}]{fujisawa2024procbenchbenchmarkmultistepreasoning}
Ippei Fujisawa, Sensho Nobe, Hiroki Seto, Rina Onda, Yoshiaki Uchida, Hiroki Ikoma, Pei-Chun Chien, and Ryota Kanai. 2024.
\newblock Procbench: Benchmark for multi-step reasoning and following procedure.

\bibitem[{Heo et~al.(2024)Heo, Xiong, Heinze-Deml, and Narain}]{heo2024douncertainity}
Juyeon Heo, Miao Xiong, Christina Heinze-Deml, and Jaya Narain. 2024.
\newblock Do {LLM}s estimate uncertainty well in instruction-following?
\newblock In \emph{Neurips Safe Generative AI Workshop 2024}.

\bibitem[{Hu et~al.(2022)Hu, yelong shen, Wallis, Allen-Zhu, Li, Wang, Wang, and Chen}]{hu2022lora}
Edward~J Hu, yelong shen, Phillip Wallis, Zeyuan Allen-Zhu, Yuanzhi Li, Shean Wang, Lu~Wang, and Weizhu Chen. 2022.
\newblock Lo{RA}: Low-rank adaptation of large language models.
\newblock In \emph{International Conference on Learning Representations}.

\bibitem[{Jang et~al.(2022)Jang, Ye, and Seo}]{jang2022can}
Joel Jang, Seonghyeon Ye, and Minjoon Seo. 2022.
\newblock Can large language models truly follow your instructions?
\newblock In \emph{NeurIPS ML Safety Workshop}.

\bibitem[{Kubiszyn and Borich(2016)}]{kubiszyn2016}
Tom Kubiszyn and Gary Borich. 2016.
\newblock Educational testing and measurement.
\newblock \emph{John Wiley and Sons, New Jersey, USA}.

\bibitem[{Lee et~al.(2025)Lee, Kim, and Jo}]{lee2025dpomcqgen}
Yooseop Lee, Suin Kim, and Yohan Jo. 2025.
\newblock Generating plausible distractors for multiple-choice questions via student choice prediction.

\bibitem[{Li et~al.(2023)Li, Zhu, Lu, and Yin}]{li2023synthetic}
Zhuoyan Li, Hangxiao Zhu, Zhuoran Lu, and Ming Yin. 2023.
\newblock Synthetic data generation with large language models for text classification: Potential and limitations.
\newblock In \emph{The 2023 Conference on Empirical Methods in Natural Language Processing}.

\bibitem[{Liang et~al.(2018)Liang, Yang, Dave, Wham, Pursel, and Giles}]{liang2018distractorrank}
Chen Liang, Xiao Yang, Neisarg Dave, Drew Wham, Bart Pursel, and C.~Lee Giles. 2018.
\newblock Distractor generation for multiple choice questions using learning to rank.
\newblock In \emph{Proceedings of the Thirteenth Workshop on Innovative Use of {NLP} for Building Educational Applications}. Association for Computational Linguistics.

\bibitem[{Lin(2004)}]{lin2004rouge}
Chin-Yew Lin. 2004.
\newblock {ROUGE}: A package for automatic evaluation of summaries.
\newblock In \emph{Text Summarization Branches Out}. Association for Computational Linguistics.

\bibitem[{Liu et~al.(2024{\natexlab{a}})Liu, Liu, Jin, Stone, and Liu}]{liu2024conflictaversegradientdescentmultitask}
Bo~Liu, Xingchao Liu, Xiaojie Jin, Peter Stone, and Qiang Liu. 2024{\natexlab{a}}.
\newblock Conflict-averse gradient descent for multi-task learning.

\bibitem[{Liu et~al.(2023)Liu, Iter, Xu, Wang, Xu, and Zhu}]{liu2023geval}
Yang Liu, Dan Iter, Yichong Xu, Shuohang Wang, Ruochen Xu, and Chenguang Zhu. 2023.
\newblock {G}-eval: {NLG} evaluation using gpt-4 with better human alignment.
\newblock In \emph{Proceedings of the 2023 Conference on Empirical Methods in Natural Language Processing}. Association for Computational Linguistics.

\bibitem[{Liu et~al.(2024{\natexlab{b}})Liu, Lu, Zhang, Liu, Guo, Yang, Blanchet, and Wang}]{liu2024provably}
Zhihan Liu, Miao Lu, Shenao Zhang, Boyi Liu, Hongyi Guo, Yingxiang Yang, Jose Blanchet, and Zhaoran Wang. 2024{\natexlab{b}}.
\newblock \href {https://openreview.net/forum?id=2cQ3lPhkeO} {Provably mitigating overoptimization in {RLHF}: Your {SFT} loss is implicitly an adversarial regularizer}.
\newblock In \emph{The Thirty-eighth Annual Conference on Neural Information Processing Systems}.

\bibitem[{Loshchilov and Hutter(2019)}]{loshchilov2018decoupled}
Ilya Loshchilov and Frank Hutter. 2019.
\newblock \href {https://openreview.net/forum?id=Bkg6RiCqY7} {Decoupled weight decay regularization}.
\newblock In \emph{International Conference on Learning Representations}.

\bibitem[{Lou et~al.(2024)Lou, Zhang, and Yin}]{lou2024largelanguagemodelinstruction}
Renze Lou, Kai Zhang, and Wenpeng Yin. 2024.
\newblock Large language model instruction following: A survey of progresses and challenges.

\bibitem[{Luo et~al.(2024)Luo, Deng, Shen, Ng, and Chua}]{luo2024cotdistractor}
Haohao Luo, Yang Deng, Ying Shen, See-Kiong Ng, and Tat-Seng Chua. 2024.
\newblock Chain-of-exemplar: Enhancing distractor generation for multimodal educational question generation.
\newblock In \emph{Proceedings of the 62nd Annual Meeting of the Association for Computational Linguistics (Volume 1: Long Papers)}. Association for Computational Linguistics.

\bibitem[{Nitko(1996)}]{anthony1996}
Anthony~J. Nitko. 1996.
\newblock Educational assessment of students.
\newblock \emph{Prentice-Hall, Iowa, USA}.

\bibitem[{Ouyang et~al.(2022)Ouyang, Wu, Jiang, Almeida, Wainwright, Mishkin, Zhang, Agarwal, Slama, Ray, Schulman, Hilton, Kelton, Miller, Simens, Askell, Welinder, Christiano, Leike, and Lowe}]{ouyang2022traininglanguagemodelsfollow}
Long Ouyang, Jeff Wu, Xu~Jiang, Diogo Almeida, Carroll~L. Wainwright, Pamela Mishkin, Chong Zhang, Sandhini Agarwal, Katarina Slama, Alex Ray, John Schulman, Jacob Hilton, Fraser Kelton, Luke Miller, Maddie Simens, Amanda Askell, Peter Welinder, Paul Christiano, Jan Leike, and Ryan Lowe. 2022.
\newblock Training language models to follow instructions with human feedback.

\bibitem[{Pal et~al.(2024)Pal, Karkhanis, Dooley, Roberts, Naidu, and White}]{pal2024smaug}
Arka Pal, Deep Karkhanis, Samuel Dooley, Manley Roberts, Siddartha Naidu, and Colin White. 2024.
\newblock Smaug: Fixing failure modes of preference optimisation with dpo-positive.
\newblock \emph{arXiv preprint arXiv:2402.13228}.

\bibitem[{Park et~al.(2024)Park, Rafailov, Ermon, and Finn}]{park2024disentanglinglengthqualitydirect}
Ryan Park, Rafael Rafailov, Stefano Ermon, and Chelsea Finn. 2024.
\newblock Disentangling length from quality in direct preference optimization.

\bibitem[{Rafailov et~al.(2023)Rafailov, Sharma, Mitchell, Manning, Ermon, and Finn}]{rafailov2023direct}
Rafael Rafailov, Archit Sharma, Eric Mitchell, Christopher~D Manning, Stefano Ermon, and Chelsea Finn. 2023.
\newblock \href {https://openreview.net/forum?id=HPuSIXJaa9} {Direct preference optimization: Your language model is secretly a reward model}.
\newblock In \emph{Thirty-seventh Conference on Neural Information Processing Systems}.

\bibitem[{Ren and Q.~Zhu(2021)}]{zhu2021cloze}
Siyu Ren and Kenny Q.~Zhu. 2021.
\newblock Knowledge-driven distractor generation for cloze-style multiple choice questions.
\newblock \emph{Proceedings of the AAAI Conference on Artificial Intelligence}.

\bibitem[{Rey and Neuh{\"a}user(2011)}]{Rey2011wilcoxonsigned}
Denise Rey and Markus Neuh{\"a}user. 2011.
\newblock \emph{Wilcoxon-Signed-Rank Test}.
\newblock Springer Berlin Heidelberg.

\bibitem[{Scarlatos et~al.(2024{\natexlab{a}})Scarlatos, Feng, Smith, Woodhead, and Lan}]{scarlatos-etal-2024-improving}
Alexander Scarlatos, Wanyong Feng, Digory Smith, Simon Woodhead, and Andrew Lan. 2024{\natexlab{a}}.
\newblock \href {https://aclanthology.org/2024.bea-1.19/} {Improving automated distractor generation for math multiple-choice questions with overgenerate-and-rank}.
\newblock In \emph{Proceedings of the 19th Workshop on Innovative Use of NLP for Building Educational Applications (BEA 2024)}, pages 222--231, Mexico City, Mexico. Association for Computational Linguistics.

\bibitem[{Scarlatos et~al.(2025)Scarlatos, Liu, Lee, Baraniuk, and Lan}]{scarlatos2025trainingllmbasedtutorsimprove}
Alexander Scarlatos, Naiming Liu, Jaewook Lee, Richard Baraniuk, and Andrew Lan. 2025.
\newblock \href {https://arxiv.org/abs/2503.06424} {Training llm-based tutors to improve student learning outcomes in dialogues}.
\newblock \emph{Preprint}, arXiv:2503.06424.

\bibitem[{Scarlatos et~al.(2024{\natexlab{b}})Scarlatos, Smith, Woodhead, and Lan}]{scarlatos2024improving}
Alexander Scarlatos, Digory Smith, Simon Woodhead, and Andrew Lan. 2024{\natexlab{b}}.
\newblock Improving the validity of automatically generated feedback via reinforcement learning.
\newblock In \emph{Artificial Intelligence in Education}, pages 280--294, Cham. Springer Nature Switzerland.

\bibitem[{Shekhar et~al.(2024)Shekhar, Singh, and Zhang}]{shekhar2024seedposelfentropyenhanced}
Shivanshu Shekhar, Shreyas Singh, and Tong Zhang. 2024.
\newblock See-dpo: Self entropy enhanced direct preference optimization.

\bibitem[{Shi et~al.(2023)Shi, Li, Zhang, Chen, and Wu}]{shi2023recon}
Guangyuan Shi, Qimai Li, Wenlong Zhang, Jiaxin Chen, and Xiao-Ming Wu. 2023.
\newblock Recon: Reducing conflicting gradients from the root for multi-task learning.
\newblock In \emph{The Eleventh International Conference on Learning Representations}.

\bibitem[{Shin et~al.(2019)Shin, Guo, and Gierl}]{shin2019topicmodelling}
Jinnie Shin, Qi~Guo, and Mark~J. Gierl. 2019.
\newblock Multiple-choice item distractor development using topic modeling approaches.
\newblock \emph{Frontiers in Psychology}, Volume 10 - 2019.

\bibitem[{Simkin and Kuechler(2005)}]{simkin2005multiple}
Mark~G Simkin and William~L Kuechler. 2005.
\newblock Multiple-choice tests and student understanding: What is the connection?
\newblock \emph{Decision Sciences Journal of Innovative Education}, 3(1):73--98.

\bibitem[{Son et~al.(2024)Son, Baek, Nam, Jeong, and Kim}]{son2024multitaskinference}
Guijin Son, SangWon Baek, Sangdae Nam, Ilgyun Jeong, and Seungone Kim. 2024.
\newblock Multi-task inference: Can large language models follow multiple instructions at once?
\newblock In \emph{Proceedings of the 62nd Annual Meeting of the Association for Computational Linguistics (Volume 1: Long Papers)}. Association for Computational Linguistics.

\bibitem[{Sonkar et~al.(2024{\natexlab{a}})Sonkar, Liu, Le, and Baraniuk}]{sonkar2024malalgoqa}
Shashank Sonkar, Naiming Liu, MyCo Le, and Richard Baraniuk. 2024{\natexlab{a}}.
\newblock Malalgoqa: Pedagogical evaluation of counterfactual reasoning in large language models and implications for ai in education.
\newblock In \emph{Findings of the Association for Computational Linguistics: EMNLP 2024}, pages 15554--15567.

\bibitem[{Sonkar et~al.(2024{\natexlab{b}})Sonkar, Ni, Chaudhary, and Baraniuk}]{sonkar2024pedagogicalalignment}
Shashank Sonkar, Kangqi Ni, Sapana Chaudhary, and Richard Baraniuk. 2024{\natexlab{b}}.
\newblock Pedagogical alignment of large language models.
\newblock In \emph{Findings of the Association for Computational Linguistics: EMNLP 2024}.

\bibitem[{Stasaski and Hearst(2017)}]{stasaski2017multipleontology}
Katherine Stasaski and Marti~A. Hearst. 2017.
\newblock Multiple choice question generation utilizing an ontology.
\newblock In \emph{Proceedings of the 12th Workshop on Innovative Use of {NLP} for Building Educational Applications}. Association for Computational Linguistics.

\bibitem[{Tan et~al.(2024)Tan, Li, Wang, Beigi, Jiang, Bhattacharjee, Karami, Li, Cheng, and Liu}]{tan2024largeannotation}
Zhen Tan, Dawei Li, Song Wang, Alimohammad Beigi, Bohan Jiang, Amrita Bhattacharjee, Mansooreh Karami, Jundong Li, Lu~Cheng, and Huan Liu. 2024.
\newblock Large language models for data annotation and synthesis: A survey.
\newblock In \emph{Proceedings of the 2024 Conference on Empirical Methods in Natural Language Processing}. Association for Computational Linguistics.

\bibitem[{Team et~al.(2024)Team, Modi, Veerubhotla, Rysbek, Huber, Wiltshire, Veprek, Gillick, Kasenberg, Ahmed, Jurenka, Cohan, She, Wilkowski, Alarakyia, McKee, Wang, Kunesch, Schaekermann, Pîslar, Joshi, Mahmoudieh, Jhun, Wiltberger, Mohamed, Agarwal, Phal, Lee, Strinopoulos, Ko, Wang, Anand, Bhoopchand, Wild, Pandya, Bar, Graham, Winnemoeller, Nagda, Kolhar, Schneider, Zhu, Chan, Yadlowsky, Sounderajah, and Assael}]{learnlmteam2024learnlmimprovinggeminilearning}
LearnLM Team, Abhinit Modi, Aditya~Srikanth Veerubhotla, Aliya Rysbek, Andrea Huber, Brett Wiltshire, Brian Veprek, Daniel Gillick, Daniel Kasenberg, Derek Ahmed, Irina Jurenka, James Cohan, Jennifer She, Julia Wilkowski, Kaiz Alarakyia, Kevin~R. McKee, Lisa Wang, Markus Kunesch, Mike Schaekermann, and 27 others. 2024.
\newblock Learnlm: Improving gemini for learning.

\bibitem[{Thomas et~al.(2025)Thomas, Borchers, Kakarla, Lin, Bhushan, Guo, Gatz, and Koedinger}]{thomas2025does}
Danielle~R Thomas, Conrad Borchers, Sanjit Kakarla, Jionghao Lin, Shambhavi Bhushan, Boyuan Guo, Erin Gatz, and Kenneth~R Koedinger. 2025.
\newblock Does multiple choice have a future in the age of generative ai? a posttest-only rct.
\newblock In \emph{Proceedings of the 15th International Learning Analytics and Knowledge Conference}, pages 494--504.

\bibitem[{Vijayakumar et~al.(2018)Vijayakumar, Cogswell, Selvaraju, Sun, Lee, Crandall, and Batra}]{vijayakumar2018diverse}
Ashwin~K Vijayakumar, Michael Cogswell, Ramprasath~R. Selvaraju, Qing Sun, Stefan Lee, David Crandall, and Dhruv Batra. 2018.
\newblock \href {https://arxiv.org/abs/1610.02424} {Diverse beam search: Decoding diverse solutions from neural sequence models}.
\newblock \emph{Preprint}, arXiv:1610.02424.

\bibitem[{Wang and Lu(2023)}]{wang2023learningmultistepreasoningsolving}
Tianduo Wang and Wei Lu. 2023.
\newblock Learning multi-step reasoning by solving arithmetic tasks.

\bibitem[{Wen et~al.(2024)Wen, Ke, Gu, Wu, Huang, Zhou, Li, Hu, Gao, Xu, Liu, Tang, Wang, and Huang}]{wen2024benchmarkingcomplexinstructionfollowing}
Bosi Wen, Pei Ke, Xiaotao Gu, Lindong Wu, Hao Huang, Jinfeng Zhou, Wenchuang Li, Binxin Hu, Wendy Gao, Jiaxing Xu, Yiming Liu, Jie Tang, Hongning Wang, and Minlie Huang. 2024.
\newblock Benchmarking complex instruction-following with multiple constraints composition.
\newblock In \emph{The Thirty-eight Conference on Neural Information Processing Systems Datasets and Benchmarks Track}.

\bibitem[{Wolf et~al.(2020)Wolf, Debut, Sanh, Chaumond, Delangue, Moi, Cistac, Rault, Louf, Funtowicz, Davison, Shleifer, von Platen, Ma, Jernite, Plu, Xu, Le~Scao, Gugger, Drame, Lhoest, and Rush}]{wolf2020transformers}
Thomas Wolf, Lysandre Debut, Victor Sanh, Julien Chaumond, Clement Delangue, Anthony Moi, Pierric Cistac, Tim Rault, Remi Louf, Morgan Funtowicz, Joe Davison, Sam Shleifer, Patrick von Platen, Clara Ma, Yacine Jernite, Julien Plu, Canwen Xu, Teven Le~Scao, Sylvain Gugger, and 3 others. 2020.
\newblock Transformers: State-of-the-art natural language processing.
\newblock In \emph{Proceedings of the 2020 Conference on Empirical Methods in Natural Language Processing: System Demonstrations}. Association for Computational Linguistics.

\bibitem[{Xu et~al.(2024)Xu, Fu, Gao, Ye, Liu, Mei, Wang, Yu, and Wu}]{xu2024isdposuperior}
Shusheng Xu, Wei Fu, Jiaxuan Gao, Wenjie Ye, Weilin Liu, Zhiyu Mei, Guangju Wang, Chao Yu, and Yi~Wu. 2024.
\newblock Is dpo superior to ppo for llm alignment? a comprehensive study.
\newblock In \emph{Proceedings of the 41st International Conference on Machine Learning}.

\bibitem[{Yan et~al.(2025)Yan, Miao, Li, YipinZhang, Xie, Deng, and Yan}]{yan2025dproperties}
Yuzi Yan, Yibo Miao, Jialian Li, YipinZhang, Jian Xie, Zhijie Deng, and Dong Yan. 2025.
\newblock \href {https://openreview.net/forum?id=9Hxdixed7p} {3d-properties: Identifying challenges in {DPO} and charting a path forward}.
\newblock In \emph{The Thirteenth International Conference on Learning Representations}.

\bibitem[{Yu et~al.(2024{\natexlab{a}})Yu, Shih, Law, Hsieh, Cheng, Ho, Lin, Hsu, and Fan}]{yu2024enhancingrag}
Han~Cheng Yu, Yu~An Shih, Kin~Man Law, KaiYu Hsieh, Yu~Chen Cheng, Hsin~Chih Ho, Zih~An Lin, Wen-Chuan Hsu, and Yao-Chung Fan. 2024{\natexlab{a}}.
\newblock Enhancing distractor generation for multiple-choice questions with retrieval augmented pretraining and knowledge graph integration.
\newblock In \emph{Findings of the Association for Computational Linguistics: ACL 2024}. Association for Computational Linguistics.

\bibitem[{Yu et~al.(2024{\natexlab{b}})Yu, Jiang, Shi, Yu, Liu, Zhang, Kwok, Li, Weller, and Liu}]{yu2024metamathbootstrapmathematicalquestions}
Longhui Yu, Weisen Jiang, Han Shi, Jincheng Yu, Zhengying Liu, Yu~Zhang, James~T. Kwok, Zhenguo Li, Adrian Weller, and Weiyang Liu. 2024{\natexlab{b}}.
\newblock Metamath: Bootstrap your own mathematical questions for large language models.

\bibitem[{Zhang et~al.(2020)Zhang, Kishore, Wu, Weinberger, and Artzi}]{zhang2020bertscoreevaluatingtextgeneration}
Tianyi Zhang, Varsha Kishore, Felix Wu, Kilian~Q. Weinberger, and Yoav Artzi. 2020.
\newblock Bertscore: Evaluating text generation with bert.

\bibitem[{Zheng et~al.(2023)Zheng, Chiang, Sheng, Zhuang, Wu, Zhuang, Lin, Li, Li, Xing, Zhang, Gonzalez, and Stoica}]{zheng2023judging}
Lianmin Zheng, Wei-Lin Chiang, Ying Sheng, Siyuan Zhuang, Zhanghao Wu, Yonghao Zhuang, Zi~Lin, Zhuohan Li, Dacheng Li, Eric Xing, Hao Zhang, Joseph~E. Gonzalez, and Ion Stoica. 2023.
\newblock Judging {LLM}-as-a-judge with {MT}-bench and chatbot arena.
\newblock In \emph{Thirty-seventh Conference on Neural Information Processing Systems Datasets and Benchmarks Track}.

\end{thebibliography}

\appendix

\section{Baselines and their Hyperparameters}
\label{adx:baselines}
We describe \projectname's baselines, as well as the hyperparameters used by \projectname and its baselines. We use MetaMath-Mistral $7$B~\citep{yu2024metamathbootstrapmathematicalquestions} as our base LLM backbone for error and distractor generation across methods. 
For memory efficiency, we quantize the model weights into 8-bit integer representation and enable gradient checkpointing throughout training. 
Our implementation utilize the HuggingFace ecosystem, specifically the \texttt{transformers} \citep{wolf2020transformers}, \texttt{peft}, and \texttt{trl} libraries for finetuning. 
We perform training on NVIDIA L40 GPUs.

\paragraph{SFT.}
For the supervised finetuning (SFT) baseline we train the base model with Low-Rank Adaptation (LoRA) modules \citep{hu2022lora}. LoRA is configured with a rank $r=128$, $\alpha=256$, and a dropout rate of 0.05. We perform SFT training for 5 epochs, with early stopping based on validation loss. We use the AdamW optimizer \citep{loshchilov2018decoupled} with a learning rate of 2e-5.
We use a batch size of $6$.

\paragraph{DPO-based Baselines.}
For all DPO training, we set the hyperparameter $\beta=0.5$ and the learning rate as 5e-6. We use a batch size of $6$.

\paragraph{DPO-GT.}
As specified in \ref{subsec:preference-pairs-from-ground-truth} we have multiple errors and distractors associated with all questions, to create preference pairs for each pair of error and distractor, we place all the non-associated sample of either in the dispreferred pair while placing the specified samples in the preferred pairs. 

\paragraph{RPO.}
For RPO (Section \ref{subsec:optimizing-dpo-sft-objectives-jointly}), we use $\lambda=0.005$ as reported by them. 
We use the default implementations of RPO as provided in the trl library.

\paragraph{DPOP}

DPO-Positive (DPOP) \cite{pal2024smaug} enhances DPO by preventing the model from merely reducing the likelihood of rejected examples where the edit distance in all pairs is large by using the SFT objective as a penalty. It introduces a constraint term to balance learning:
\begin{equation}
    L_{DPOP} = L_{DPO} - \lambda \cdot \max(0, \log \frac{\pi_{ref}(y_w|x)}{\pi_\theta(y_w|x)}).
\end{equation}
Here, we use $\lambda=0.1$.

\paragraph{\projectname (Synthetic Data Generation).}
For the \projectname prefrence pairs (in Section \ref{subsec:mining_preference_pairs}) we generate $3$ errors and distractors for each epoch of training to create negative preference samples, while considering the ground truth errors and distractors as the positive preference samples. 
We consider the top-$k$ completions returned by beam search to get a set of $\hat{e}_i$ which augments the set of dispreferred responses further.
We note that for all DPO training we use the SFT trained model as a warm start as with previous literature \cite{rafailov2023direct}. 

\paragraph{\projectname (Per-epoch and Per-batch Regularization).}
With the per-epoch and per-batch modes of \projectname (Section \ref{subsec:alternating_sft_dpo_introduction}), we use the learning rate of 5e-6 for both DPO and SFT. 
For the per-epoch setting we perform one entire epoch of SFT after one epoch of DPO.
Whereas for the per-batch setting if we run out of SFT batches while DPO training hasn't finished we rollback to the beginning of the SFT training data. 

\section{LLM-as-a-judge}
\label{adx:llm-as-a-judge}
To assess whether two error explanations express the same underlying misconception, we use \texttt{GPT-4o-mini} as an automated judge. 
The model is provided with the question, distractor, and two error explanations, and asked to determine whether they are \textit{mathematically equivalent} (Table~\ref{tbl:llm-as-a-judge-prompt}), that is, whether they arise from the same conceptual misunderstanding, regardless of wording. 
Below, we present an example of the prompt used in this evaluation.

\begin{table*}
\begin{center}
\begin{minipage}{\textwidth}
\begin{promptbox}
\textbf{System Prompt.}

You are a math education expert.

Given a question and a distractor (an incorrect student answer), determine whether two error descriptions are \textit{mathematically equivalent}.\\

\textbf{Definitions.}
\begin{itemize}[noitemsep]
    \item An incorrect answer or distractor is a plausible but incorrect answer choice to the specified question.
    \item An error explanation or error is the misconception a student might make that leads them to choosing the specified distractor.
    \item Two error explanations are \textit{mathematically equivalent} if they stem from the same core misunderstanding, regardless of wording.
\end{itemize}

Your response should include a brief justification (1–2 sentences) for whether the errors reflect the same or different misconceptions.

Always conclude with: ``\textbf{Answer: Equivalent} or \textbf{Answer: Not Equivalent}''.\\

\textbf{Question and Metadata.}

The question is: \texttt{<Question>} \\
The question topic is: \texttt{<Topic>} \\
The question concept is: \texttt{<Concept>} \\
The solution is: \texttt{<Solution from question to Correct Answer>} \\
The correct answer is: \texttt{<Correct Answer>}\\ 

\vspace{0.5em}
Distractor (incorrect answer): \texttt{<Ground Truth Distractor>} \\

Error explanation 1: \texttt{<Ground Truth Error>} \\
Error explanation 2: \texttt{<Generated Error>}
\end{promptbox}
\end{minipage} 
\caption{
System prompt used to evaluate the mathematical equivalence of error explanations for a given distractor. 
The prompt positions the model as a math education expert tasked with identifying whether two misconceptions arise from the same underlying error.
}
\label{tbl:llm-as-a-judge-prompt}
\end{center}
\end{table*}

This template was used for all pairwise comparisons of error explanations in the LLM-as-a-Judge evaluation.

\section{Error Analysis}
\label{adx:error-analysis}

While \projectname generally produces more specific and grounded error explanations, Table~\ref{tab:appendix-error-comparison} also reveals some notable limitations. 
In the cube root example, the explanation “Has multiplied by the root power” reflects a plausible arithmetic confusion but doesn't clearly connect to the distractor value of 64, which results from cubing rather than misunderstanding cube roots. 
Similarly, in the number ordering case, the generated error implies digit-level misordering but lacks clarity on how this leads specifically to choosing “Only Katie.” 
These examples suggest that while \projectname often captures fine-grained misconceptions, it can occasionally overgeneralize or introduce speculative reasoning not fully aligned with the distractor. 
This underscores the need for further refinement to ensure tighter alignment between the error explanation and the underlying choice.

\begin{table*}[t]
\centering
\rowcolors{2}{white}{gray!20}
\begin{tabularx}{\textwidth}{@{}l|X|X@{}}
\toprule
\textbf{Field} & \textbf{Cube Root} & \textbf{Indices, Powers and Roots}\\
\midrule
{Question} & 
\( \sqrt[3]{8} = \, ? \) &
\( 3.52+2.75= \)
 \\
{Distractor} & $64$ & $5.27$ \\
{Correct Answer} & $2$ & $6.27$\\
{SFT Error Explanation} &
Divides by the order of the root. &
Does not understand place value within a number. \\
{\projectname Error Explanation} &
Has multiplied by the root power. &
When adding decimals with a different number of decimal places, lines up the digits incorrectly. \\
\bottomrule
\end{tabularx}
\caption{Comparison of error explanations for two different math topics. Examples show that \projectname also has some failure modes, discussed in greater depth in Section~\ref{sec:qualitative_analysis}.}
\label{tab:appendix-error-comparison}
\end{table*}

\section{Comparing Errors across \projectname and its Baselines}
\label{adx:comparing-errors-across-all}
Table~\ref{tbl:comparing_errors_across_all_table} illustrates how different training methods produce qualitatively distinct reasoning errors across representative math questions. We observe a clear progression in the nature of these errors, reflecting the underlying supervision strategies. Models trained with SFT often generate surface-level mistakes indicative of limited conceptual understanding. In contrast, DiVERT tends to produce more structured but still incorrect procedural reasoning. Errors from DPO-GT reveal partial application of mathematical heuristics, suggesting more sophisticated—though still flawed—mental models. Finally, \projectname models (both per batch and per epoch) consistently produce errors that resemble common student misconceptions, such as overgeneralizing valid procedures or subtly misapplying familiar rules. This progression supports our claim that \projectname encourages more pedagogically meaningful error patterns, aligning closely with authentic human reasoning.

\renewcommand{\arraystretch}{1.5}
\begin{table*}[ht!]
    \centering
    \small
    \begin{tabular}{l|p{6cm}|p{6cm}}
    \toprule
     & \textbf{Improper Fraction Conversion} & \textbf{Gradient of a Line} \\
    \midrule
    Question & Convert this into an improper fraction: \(4 \frac{2}{3}\) & What is the gradient of this line? \(5x + 3y = 15\) \\
    Correct Answer & \(\frac{14}{3}\) & \(-\frac{5}{3}\) \\
    Ground-truth Distractor & \(\frac{12}{3}\) & \(\frac{5}{3}\) \\
    Ground-truth Error & Forgetting to add the numerator to the whole part. & Applying the same operation instead of the inverse when solving. \\
    \midrule
    SFT & Does not add the whole to the numerator when converting a mixed number to an improper fraction. & Believes a downward line has a positive gradient. \\
    \midrule
    DiVERT & Thinks you add the number of wholes to the numerator when converting a mixed number to an improper fraction. & When solving an equation, uses the same operation rather than the inverse. \\
    \midrule
    DPO-GT + Per batch & Does not include the whole amount when converting a mixed number to an improper fraction. & Believes the gradient of a line is given by the coefficient of \(x\), even when the equation is not in the form \(y = mx + c\). \\
    \midrule
    \projectname + Per batch & Thinks you add the number of wholes to the numerator when converting a mixed number to an improper fraction. & When solving an equation, uses the same operation rather than the inverse. \\
    \midrule
    DPO-GT + Per epoch  & Thinks you can ignore the whole amount when converting a mixed number to an improper fraction. & When finding the gradient from the equation of a line in the form \(ax + by = c\), believes \(b/a\) is the gradient. \\
    \midrule
    \projectname + Per epoch & Thinks you add the number of wholes to the numerator when converting a mixed number to an improper fraction. & When finding the gradient from the equation of a line in the form \(ax + by = c\), believes \(b/a\) is the gradient. \\
    \bottomrule
    \end{tabular}
    \caption{
    Comparison of typical errors generated by each method for two representative math questions. \\
}
\label{tbl:comparing_errors_across_all_table}
\end{table*}
\renewcommand{\arraystretch}{1}

\section{Example MCQs from Real-world Math
MCQ Dataset}
\label{adx:examples-from-eedi}
We show example MCQs from the dataset in Ta-
ble \ref{tbl:examples-from-eedi}.

\begin{table*}[ht!]
\begin{tabular}{@{} l| p{0.8\linewidth} @{}}
\toprule
\rowcolor{gray!20}
Question stem & Add brackets to this calculation to make the answer \( 7 \). \( 16-6+4 \div 2 \) \\
\midrule
Topic & BIDMAS \\
Concept & Insert brackets to make a calculation correct \\
Solution & Inside the bracket we work left to right, so we get \(14 \div 2\) which is 7. \\
Correct answer & \( (16-6+4) \div 2 \) \\
\midrule
Distractor 1 & \( 16-(6+4)\div2 \) \\
\midrule
Error 1 & With order of operations brackets are done first, then division is done before subtraction. This would give us \( 16 - 10 \div 2 = 16 - 5 = 11 \) NOT 7. \\
\midrule
Distractor 2 & \( (16-6)+\frac{4}{2} \) \\
\midrule
Error 2 & With order of operations brackets are done first, then division is done before subtraction. This would give us \( 10 + 4 \div 2 = 10 + 2 = 12 \) NOT 7. \\
\midrule
Distractor 3 & \( 16-6+(\frac{4}{2}) \) \\
\midrule
Error 3 & With order of operations brackets are done first, then division is done before subtraction. Putting the brackets around the division, will not change the order. \( 16 - 6 + (4 \div 2) = 16 - 6 + 2 = 12 \) NOT 7. \\
\midrule \midrule
\rowcolor{gray!20}
Question stem & Which of the following answers gives the correct solutions to the quadratic expression below? \( (x+2)(x-7)=0 \) \\
\midrule
Topic & Algebra \\
Concept & Solve quadratic equations using factorisation in the form (x + a)(x + b)
 \\
Solution & Setting each bracket equal to 0 we have x + 2 = 0 and x - 7 = 0. This tells us that x = -2 and x = 7. \\
Correct answer & \( x=-2, x=7 \) \\
\midrule
Distractor 1 & \( x=2,x=-7 \) \\
\midrule
Error 1 & Believes the solutions of a quadratic equation are the constants in the factorised form \\
\midrule
Distractor 2 & \( x=2,x=7 \) \\
\midrule
Error 2 & Believes the solutions of a quadratic equation are the absolute values of the constants in the factorised form
 \\
\midrule
Distractor 3 & \(x=-2,x=-7 \) \\ 
\midrule
Error 3 & Believes the solutions of a quadratic equation are the negative of the absolute values of the constants in the factorised form
 \\
\bottomrule
\end{tabular}
\caption{Example MCQs from the real-world math MCQ dataset.}
\label{tbl:examples-from-eedi}
\end{table*}

\section{Human Analysis Instructions}
\label{adx:human_analysis_instructions}
To evaluate the consistency of error explanations with corresponding distractor choices in multiple-choice math questions, we provided annotators with detailed guidelines, shown in Table~\ref{tbl:human_analysis_instructions}. Annotators were instructed to examine each question item, which included a correct answer, a step-by-step solution, a distractor (incorrect answer), and an explanation for why a student might choose that distractor.

Annotators were asked to judge whether the explanation was:

    \begin{itemize}
        \item Yes: Clearly consistent with the distractor and plausibly explains the student error.
        \item Partially: Somewhat consistent, but vague, generic, or only loosely related to the distractor.
        \item No: Inconsistent or misleading; does not plausibly explain the choice of the distractor.
    \end{itemize}
    
The instructions included concrete examples for each category to help calibrate judgment and ensure consistent annotation. These annotations were later used to analyze the quality of generated error explanations.

\begin{table*}
\begin{center}
\begin{minipage}{\textwidth}
\begin{promptbox}
In this task, you'll evaluate error explanations for student errors in math multiple-choice questions. For each item, you’ll see:

\begin{enumerate}
    \item    The \textbf{question}
    \item    The \textbf{correct answer} choice
    \item    A \textbf{solution} which shows how a student can reach the correct answer choice
    \item     A \textbf{distractor} (an incorrect answer choice)
    \item     An \textbf{error} explanation describing why a student might choose the distractor

\end{enumerate}

\textbf{Your Task}

Annotate if each error explanation is consistent with the distractor (mark Yes), is generic, vague, or partially consistent (mark Partially) or has nothing to do with the distractor, or is misleading (mark No).

Use your best judgment when assigning ratings. Some examples are:

\vspace{7px}

Example 1 (Marking \textbf{Yes}):

Question: Add brackets to this calculation to make the answer \( 7 \). \( 16-6+4 \div 2 \)

\vspace{7px}

Correct Answer:\( (16-6+4) \div 2 \)
\vspace{2px}

Solution: Inside the bracket we work left to right, so we get 14 ÷ 2 which is 7.
\vspace{2px}

Distractor: $16-(6+4)\div2$
\vspace{2px}

Error: Carries out operations from left to right regardless of priority order.

Mark \textbf{Yes}

\vspace{7px}

Example 2 (Marking \textbf{Partially}):

Question:  \( \frac{3}{7} \) of a group of students are boys. What would be a possible ratio of boys to girls?

\vspace{7px}

Correct Answer: $3: 4$
\vspace{2px}

Solution: For every 7 students, 3 are boys and 4 are girls. The ratio is then 3:4.
\vspace{2px}

Distractor: $3:7$
\vspace{2px}

Error: Uses the denominator when converting from fractions to ratio, rather than numerator.

Mark \textbf{Partially}

\vspace{7px}

Example 3 (Marking \textbf{No}):

Question:When \( h=5 \)
\(
h^{2}=
\)

\vspace{7px}

Correct Answer: \( 25 \)
\vspace{2px}

Solution: If $h = 5$, $h^2=h \times h=5 \times 5=25$.

\vspace{2px}
Distractor: $7$

\vspace{2px}
Error: Multiplies by the index.

Mark \textbf{No}
\end{promptbox}
\end{minipage} 
\caption{
Instructions provided to human annotators used to evaluate the consistency of error explanations for a given distractor. 
}
\label{tbl:human_analysis_instructions}
\end{center}
\end{table*}

\end{document}